\documentclass[a4paper,12pt]{spieman}  
\usepackage{amsmath,amsfonts,amssymb}
\usepackage{graphicx}
\usepackage{multirow}
\usepackage{comment}
\usepackage{balance}
\usepackage{setspace}
\usepackage{tocloft}

\title{A Benchmark for  License Plate Character Segmentation}

\author[a]{Gabriel Resende Gon\c{c}alves}
\author[b]{Sirlene Pio Gomes da Silva}
\author[c]{David Menotti}
\author[a]{William Robson Schwartz}
\affil[a]{Universidade Federal de Minas Gerais, Smart Surveillance Interest Group, Department of Computer Science, Belo Horizonte, Brazil, 31270-010}
\affil[b]{Universidade Federal de Ouro Preto, Computing Department, Ouro Preto, Brazil, 35400-000}
\affil[c]{Universidade Federal do Paran\'a, Department of Informatics, Curitiba, Brazil, 81531-980}

\cftpagenumbersoff{figure}
\cftpagenumbersoff{table} 
\begin{document} 
\sloppy
\maketitle

\begin{abstract}
Automatic License Plate Recognition (ALPR) has been the focus of many researches in the past years. In general, ALPR is divided into the following problems: detection of on-track vehicles, license plates detection, segmention of license plate characters and optical character recognition (OCR). Even though commercial solutions are available for controlled acquisition conditions, e.g., the entrance of a parking lot, ALPR is still an open problem when dealing with data acquired from uncontrolled environments, such as roads and highways when relying only on imaging sensors. Due to the multiple orientations and scales of the license plates captured by the camera, a very challenging task of the ALPR is the License Plate Character Segmentation (LPCS) step, which effectiveness is required to be (near) optimal to achieve a high recognition rate by the OCR. To tackle the LPCS problem, this work proposes a novel benchmark composed of a dataset designed to focus specifically on the character segmentation step of the ALPR within an evaluation protocol. Furthermore, we propose the Jaccard-Centroid coefficient, a new evaluation measure more suitable than the Jaccard coefficient regarding the location of the bounding box within the ground-truth annotation. The dataset is composed of 2,000 Brazilian license plates consisting of 14,000 alphanumeric symbols and their corresponding bounding box annotations. We also present a new straightforward approach to perform LPCS efficiently. Finally, we provide an experimental evaluation for the dataset based on five LPCS approaches and demonstrate the importance of character segmentation for achieving an accurate OCR.
\end{abstract}


{\noindent \footnotesize\textbf{*}Gabriel Resende Gon\c{c}alves, \linkable{gabrielrg@dcc.ufmg.br} }

\begin{spacing}{2}   

\section{Introduction}
\label{sec:intro}
Over the years, many researchers have focused their attention on the automatic identification of vehicles on the road, task known as Automatic License Plate Recognition (ALPR). Tackling this problem is fundamental to perform very important tasks in an automatic way, such as traffic speed control, identification of stolen vehicles, vehicle access control in private spaces and toll collection. Currently, such tasks are only successfully on controlled environments~\cite{du2013automatic}. Therefore, many companies and government departments are interested on improving their systems of traffic monitoring which justifies the need to develop an accurate and efficient approach to ALPR on uncontrolled environments.

In general, the ALPR approaches are divided into multiples subtasks that are executed in the following sequence \cite{du2013automatic}: (1)~vehicle detection in a image sequence (video); (2)~license plate detection; (3)~character segmentation; and (4)~character recognition. Nevertheless, not every work performs all of these several steps. For instance, some of them~\cite{prates2014brazilian,mendes2011,sarfraz2003saudi} try to detect the license plates in the entire scene instead of detecting the vehicle first.

\begin{figure}[!t]
\centering
\includegraphics[width=0.6\linewidth]{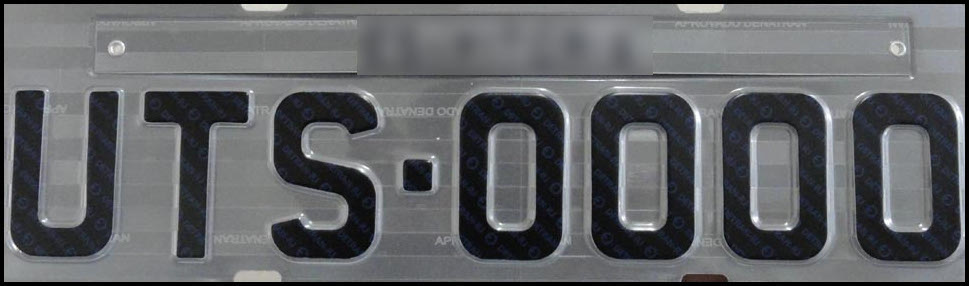}
\caption{Example of the Brazilian license plate standard. It is composed by two rows: in the first one, the acronym of state followed by its origin city (blurred in the image); in the second row, under the first one, there are three letters one blank space or hyphen and four digits to identify the vehicle.}
 \label{intro:BLPexample} 
\end{figure}

License plate character segmentation (LPCS) is a very important subtask of ALPR. A precise segmentation is essential to achieve outstanding results (accuracy near 100\%) on the next ALPR step, the Optical Character Recognition (OCR)~\cite{menotti2014vehicle,araujo2013segmenting}, as one can see on the performed experiments. Hence, in this work, we propose a new iterative technique to perform LPCS. Nonetheless, the LPCS methods are evaluated considering a large number of different datasets (not always publicly available) and a myriad of evaluation metrics, making their comparison, a very hard work.

To have a common evaluation environment for the license plate character segmentation, this work proposes a protocol for benchmarking LCPS approaches. A benchmark is a process to compare multiple approaches under some specific environment that must be fixed for all executions of the process. Specifically, our proposed benchmark is composed of (i) a new public dataset (to the best of our knowledge, this is the first dataset focused only on the license plate character segmentation task) containing 2,000 images of Brazilian license plates acquired by a digital camera at the Federal University of Minas Gerais campus, containing a total of $14,000$ characters with bounding box annotations once the Brazilian license plate standard is composed of seven characters (three letters, a hyphen and four digits), as illustrated in Figure~\ref{intro:BLPexample}; (ii) a novel evaluation measure which is more suitable to the LPCS problem than the commonly employed Jaccard coefficient; and (iii) a comparative study of many LCPS techniques using the novel dataset and measure.

The main contributions of this study can be pointed out as follows:
\begin{itemize}
	\item A new dataset with 14,000 license plate characters, ground truth annotations and an evaluation protocol to assess the quality of license plate character segmentation techniques\footnote{This dataset will made publicly available to the research community.};
	\item A new evaluation measure, the \emph{Jaccard-Centroid}, which is more suitable to the segmentation problem than the original Jaccard coefficient;
	\item A straightforward iterative approach to perform LPCS;
	\item A comparative evaluation of license plate character segmentation methods including our proposed approach.
\end{itemize}

We consider three baseline segmentation approaches using different techniques~\cite{nomura2009morphological,shapiro2004multinational,kavallieratou2006adaptive} and a straightforward technique using only the geometrical information, i.e., the size of the license plate and its characters. The experimental results demonstrate the importance of the segmentation approaches to achieve an accurate optical character recognition. We also evaluate the number of license plates well-segmented by the techniques and some cases where they were not satisfactory, indicating that no evaluated method was able to achieve very accurate results, demonstrating that the dataset is challenging to the LPCS, being suitable and interesting for research purposes.

The remainder of this paper is organized as follows. Section~\ref{sec:related} describes some related works related including the techniques used as baselines to our benchmark. In Section~\ref{sec:benchmark}, we present our new proposed technique and the protocol to our benchmark divided in three parts. Section~\ref{sec:experiments} presents the experiments conducted to evaluate the benchmark as well as the achieved results are described. Finally, in Section \ref{sec:conclusion}, we pointed out the conclusions obtained.

\section{Related Work}
\label{sec:related}
Many researchers have investigated automatic license plate recognition and its subtasks. Since this work focuses on the license plate character segmentation (LPCS) and on evaluation datasets, this section briefly reviews related works. 
We refer the reader to works such as~\cite{anagnostopoulos2008license,du2013automatic} for further information on the ALPR problem. The remaining of this section focuses on two aspects.  First, we review works related to techniques of character segmentation. Then, we present works that propose character segmentation evaluation datasets used in different contexts. Finally, we present the works describing the techniques used as baselines in this work

\subsection{License Plate Character Segmentation}

Besides license plates, there are works that propose character segmentation on various contexts. Some of them focus on handwritten text segmentation, such as in Jun et al.~\cite{tan2012new} that proposed two methods using non-linear clustering methods, and in Ciresan et al.~\cite{ciresan2011convolutional}, that uses seven convolutional networks executing in GPUs (Graphics Processing Unit) for this purpose. In Roy et al.~\cite{roy2012multi} and Neumann \& Matas~\cite{neumann2012real}, the authors propose character segmentation methods to handle digital documents and real scenes. The main goal of those works is to present an approach to segment characters on other contexts. However, they do not present promising effectiveness on license plate segmentation because they do not explore the contextual information found in licenses plates.

The LPCS can be seen as a challenging task that must be performed by ALPR systems once the acquisition of plate images usually is affected by problems such as skew, shadows, perspective projection and blurring. In an attempt to reduce that, the majority of works segment characters in manually cropped images to evaluate the effectiveness of the recognition methods. However, the license plates must be cropped automatically when applied on real applications, tackling the aforementioned problems~\cite{wang2013novel,kahraman2003license}.

Du et al.~\cite{du2013automatic} classify the license plate character segmentation techniques into five main categories: based on pixel connectivity, pixel projection, prior knowledge of the characters, characters contours and based on the combination of these features.

Many works employ approaches based on Connected Component Analysis (CCA) and pixel projection techniques to tackle the LPCS problem. A study of the effectiveness of the CCA technique concluded that vertical projection can segment characters effectively~\cite{xia2011study}.  The segmentation approach proposed by Shapiro \& Gluhchev~\cite{shapiro2004multinational} utilizes an adaptive iterative thresholding approach to binarize the image and then segment the plate characters by employing a connected component analysis. Jagannathan et al.~\cite{jagannathan2013license} proposed a method that uses ten samples of the same plate, binarize them, select the best one and then segment it using vertical projections. The work described in Soumya et al.~\cite{soumya2014license} segments characters counting the black pixels in the horizontal and vertical direction of each license plate region. Araujo et al.~\cite{araujo2013segmenting} proposed a technique to segment the characters using CCA and evaluate it in databases manually and automatically cropped.

Some works focus on other techniques to segment the license plates characters. For instance, Kahraman et al.~\cite{kahraman2003license} employed a Gabor transform and vector quantization to LPCS.
In Xing-lin \& Yun-lou~\cite{xing2012new}, the authors proposed a technique to segment the characters using prior knowledge regarding the shape and the license plate font considering English (Latin) and Chinese characters. In addition,  there are works that employ additional techniques to improve the quality of the results. For instance, Wang et al.~\cite{wang2013novel} use two segmentation techniques in sequence and Chuang et al.~\cite{chuang2014vehicle} applies super-resolution techniques.

Some statistical-based and machine learning approaches also have been employed on LPCS. For instance, while Fan et al.~\cite{fan2012license} used likelihood maximization to find the best parameters values of the license plate features and its characters and Franc at al.~\cite{franc2005license} proposed a technique using Hidden Markov Models to create a relationship between the license plate input and the correct segmentation of its characters. Nagare~\cite{nagare2011license} and Guo et al~\cite{guo2008license} employed supervised machine learning techniques to aid the character segmentation phase of the ALPR.

\subsection{Evaluation Datasets}

There are works proposing datasets to evaluate several aspects of text recognition and document analysis.  For instance, Antonacopoulos et al.~\cite{antonacopoulos2009realistic} proposed a dataset to evaluate techniques of document layout analysis. That dataset contains $1,240$ images from websites, newspaper pages, magazines pages. The UNIPEN dataset was proposed in Guyon et al.~\cite{guyon1994unipen} and  is composed by over $23,000$ images of words and handwritten characters. Yao et al.~\cite{yao2012detecting} proposed a dataset of real images to evaluate approaches to perform text detection containing 500 real images in various sizes.

There are also datasets to evaluate ALPR approaches. Two UIUC datasets were proposed in Agarwal \& Roth~\cite{agarwal2002learning} and Agarwal et al.~\cite{agarwal2004learning} composed of $170$ cars to single-scale approaches and $108$ cars to multi-scale approaches, respectively. The Caltech dataset~\cite{caltech2001dataset} provides $526$ rear images of cars and is commonly used to vehicle recognition and license plates detection. In Krause et al.~\cite{krause2013collecting}, the authors collected $16,185$ images of cars from the websites such as Flickr, Google and Bing. The BIT Dataset~\cite{dong2014vehicle} contains images of $900$ cars and aims at evaluating  techniques to recognize the vehicle type. In addition, there are other  datasets~\cite{ferencz2004learning,ozuysal2009pose} designed to evaluate tasks such as vehicle pose estimation and vehicle detection. 

To the best of our knowledge, there is no publicly available dataset to evaluate specifically license plate characters segmentation techniques, which emphasizes the contribution of this work. The proposed dataset will make easier to compare the LPCS methods since it provides a evaluation protocol in which the images are separated in training, validation and testing, allowing a fair comparison between different methods.

\subsection{Baseline Approaches}

This subsection describes four LPCS techniques chosen to be our baselines. We considered approaches based on three methods available in the literature, in which the first aims at improving the quality of degraded images of words~\cite{nomura2009morphological} and counting the blacks pixels of the image (Section~\ref{sec:pca}); the second performs segmentation by find connected components in a binarized license plate~\cite{shapiro2004multinational} (Section~\ref{sec:ccla}); and the third performs a pixel counting as well as the first technique, but utilize a license plate binarized by a method of Iterative Global Threshold (IGT)~\cite{kavallieratou2006adaptive} (Section~\ref{sec:igt}. In addition, a simple technique that employs prior knowledge regarding the license plate layout and its number of characters was used as a fourth approach and will be described in Section~\ref{sec:experiments}.

\subsubsection{SL*L Approach}
\label{sec:pca}

\begin{figure}[!t]
	\centering
	\includegraphics[width=0.45\linewidth]{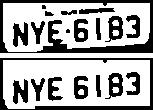}
	\caption{At the top there is an example of an image binarized without the SL*L preprocessing and at the bottom an image binarized using the SL*L processing method.}
	\label{fig:method1}
\end{figure}

This approach takes into account a very specific preprocessing method called Shadow Location and Lightening (SL*L) to improve the quality of degraded images containing text.  This method consists of a sequence of mathematical morphological operations applied to the image to locate the shadow regions on the image and lightening them to remove the noise for the final thresholding process (binarization) by the Otsu method~\cite{otsu1975threshold}. Figure~\ref{fig:method1} shows the difference when the SL*L preprocessing approach is employed.

\begin{figure}[!t]
	\centering
	\includegraphics[width=0.45\linewidth]{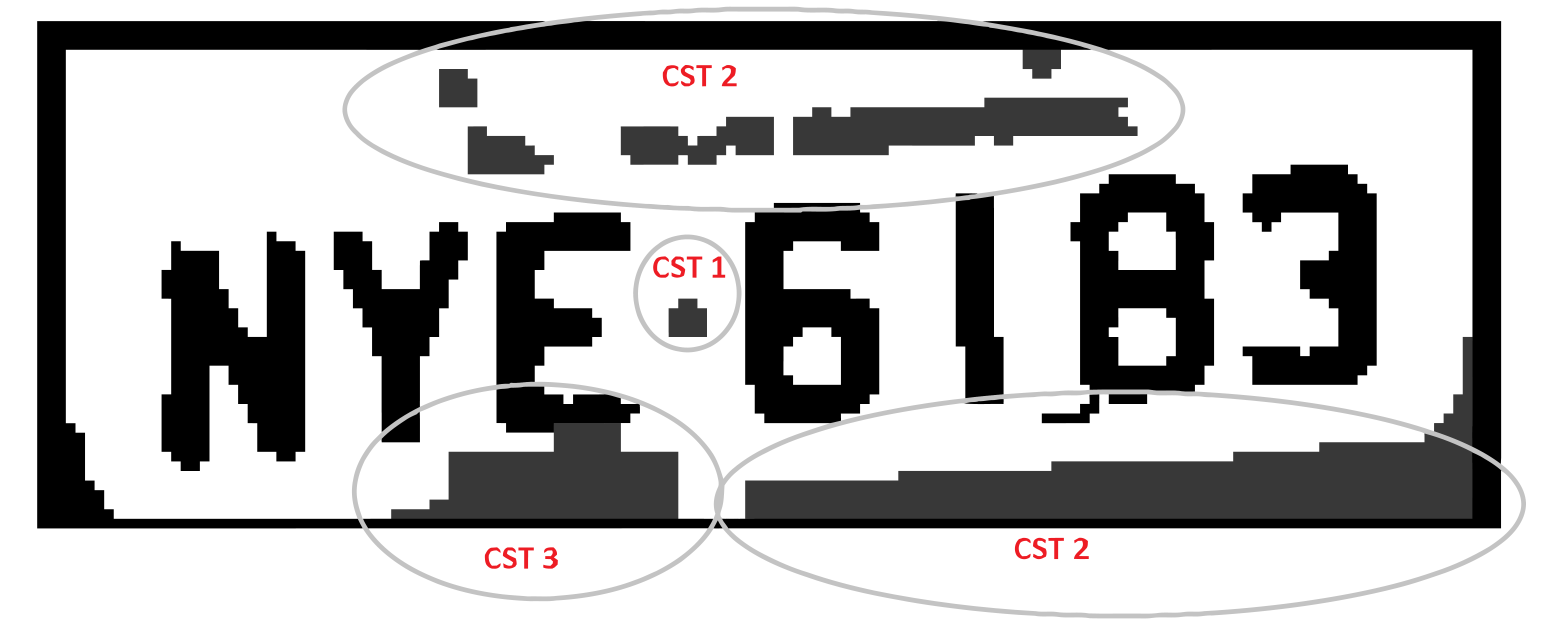}
	\caption{The three shadow types (CST1, CST2, and CST3 - in gray) that the approach must reduce.}
	\label{fig:method2}
\end{figure}

The approach begins with a binarization of the image and application of a thickening operation. Then, it locates regions with three types of shadows to reduce their effect. These types are named Critical Shadow Type 1 (CST1), 2 (CST2), and 3 (CST3). The CST1 is the shadow that can occur between the characters, CST2 is the shadow that does not occur between two characters and does not touch them, and the CST3 is the shadow that does not occur between two characters but touches one. Figure~\ref{fig:method2} illustrates these three types on a license plate converted to grayscale.

The CSTs are detected using the pruning algorithm based on 

\begin{equation}
 	S \ominus B = \left ( \bigcup_i^n (S * B^i) \oplus H \right ) \cap X,
	\label{metho:equation1}
\end{equation}

\noindent where $X$ is the binary image, $S$ is the skeleton of $X$, $B$ and $H$ are structuring elements and $n$ is the number of iterations; the operation * denotes hit-or-miss transform and $\oplus$ is the dilatation operation~\cite{serra1986introduction}.
After applying this pruning process, the image presents an enclosing boundary surrounding the shadowed regions, highlighting these regions such that a noiseless image is obtained. 
Finally, to perform the segmentation, the approach binarizes the image using a global thresholding technique and count the white pixels on both directions in order to find the segmentation points.

\subsubsection{CCL Approach}
\label{sec:ccla}

The approach proposed by Shapiro \& Gluhchev~\cite{shapiro2004multinational} is straightforward.
An adaptive thresholding is performed in the image using the Otsu method, followed by connected components labeling and then a greedy selection process is performed to chose the best characters based on their size. Each connect component is analyzed based on its height with respect to the license plate in a similar way as the iterative approach described in Section~\ref{subsec:proposed}.

Based on the real proportion of the height of a character regarding the height of a Brazilian license plate ($45\%$), we use the proportion height range of $[40\%,50\%]$  to accept a connect component as a character.

\subsubsection{IGT Approach}
\label{sec:igt}

In this approach, the segmentation is performed in binarized images of the license plates. The binarization was originally proposed in Kavallieratou~\cite{kavallieratou2005binarization} and extended in Kavallieratou \& Stathis~\cite{kavallieratou2006adaptive} in which it receive the name of Iterative Global Thresholding (IGT). The core of the algorithm consists in two steps that are applied iteratively. We consider the extended approach in this work. In the first step of IGT, we apply the following formula to all pixels of the image

\begin{equation}
 	I_f(x,y) = 1 - T_i + I(x,y)
	\label{metho:equation2}
\end{equation}

\noindent where $I_f(x,y)$ stands for the pixel $(x,y)$ at the end of the first step, $T_i$ is the average pixel of the image and $I(x,y)$ is the input image. The second step is a histogram equalization given by the formula

\begin{equation}
 	I_l(x, y) = 1 - \frac{1-I_f(x,y)}{{1-E_i}}
	\label{metho:equation3}
\end{equation}

\noindent where $I_l(x,y)$ is the image at the end of each iteration and $E_i$ is the minimum pixel of the image. In this case, the algorithm breaks when the average pixel have few variation between two consecutive iterations.

In the extension, the authors proposed an hybrid approach based on the same technique applied both globally and locally on the image. In this case, after apply the IGT, a window is slided over the image searching for areas with remaining noises. Whenever a noised area is discovered, the IGT is re-applied only in that specific region. The area is considered noised if satisfy the following condition

\begin{equation}
 	f(S) > m + (k*s)
	\label{metho:equation4}
\end{equation}

\noindent where $f(S)$ stands for the amount of black pixels in the area $S$, $m$ and $s$ are, respectivelly, the average and standard deviation of black pixels for all areas of the image, and $k$ is a parameter constant (normally 1). After the binarization, the characters bounding boxes are defined by counting the black pixels on both directions similarly to the approach described in Section~\ref{sec:pca}.

\section{Proposed Benchmark}
\label{sec:benchmark}
This section describes a new straightforward technique to iteratively perform LPCS  (Subsection~\ref{subsec:proposed}) and the protocol for our benchmarking LPCS approaches. This benchmark is composed of (i) a new public dataset (to the best of our knowledge, this is the first dataset focused only on the license plate character segmentation task) (Subsection~\ref{subsec:dataset}); (ii) and a novel evaluation measure to evaluate segmentation approaches (Subsection~\ref{subsec:measures}).

\subsection{Proposed Iterative LPCS Technique}
\label{subsec:proposed}
We developed a straightforward iterative technique to perform LPCS on real scenarios. It is composed by two steps: (i) binarization and (ii) find connected components. A similar idea was used in Matas \& Zimmermann~\cite{matas2005unconstrained} to find threshold for cars and license plates images.

In this  approach, instead of using a single threshold to perform license plate binarization using the Otsu method~\cite{gonzalez2009digital}, we consider a set of different values. Starting from a threshold equals 10, we binarize the image as we increase this threshold until we have the number of connected components equals to the number of license plate characters. By doing this, we are trying to avoid the problem where two adjacent characters are touching each other due to some noise pixel, because a binarization starting from small thresholds tends to set most pixels to the maximum value, resulting in fewer white noises connecting two adjacent characters. At each iteration, we discard connected components that are too large and too small to be a character according to the width and the area of the component. We also merge all connected components that overlaps on the x-axis.

\begin{figure}[!t]
  \centering
  \includegraphics[width=0.7\columnwidth]{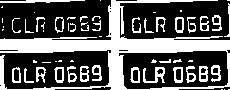}
  \caption{Samples of the license plate considering different thresholds, 1 and 10 on the top images and 20 and 30 on the bottom images.}
    \label{proposed/figure2}
\end{figure}

Figure~\ref{proposed/figure2} illustrates the binarization process. Note that when the threshold is too small, we tend to have more connected components due to sliced characters and when the threshold is too large, we have few connected components due to noises that create touching characters.

\begin{figure}[!t]
  \centering
  \includegraphics[width=0.8\columnwidth]{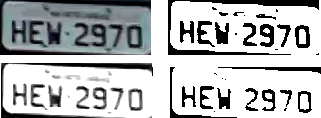}
  \caption{Comparision between images. Original image (top left), binarized by Otsu method (top right), binarized by IGT method (bottom left) and binarized by the proposed iterative approach (bottom right).}
    \label{proposed/figure3}
\end{figure}

In addition, Figure~\ref{proposed/figure3} shows a comparision between the segmentation obtained by the original image, Otsu method~\cite{gonzalez2009digital}, Iterative Global Thresholding method~\cite{kavallieratou2006adaptive} which is used in one of the described baselines used in our benchmark and our proposed iterative technique.

\subsection{Proposed Dataset}
\label{subsec:dataset}
To be able to evaluate techniques of license plate character segmentation, we compiled a large set of images of on-track vehicles and their license plates into a novel dataset. This dataset, called \emph{SSIG-SegPlate}\footnote{The SSIG-SegPlate dataset is publicly available to the research community at \url{http://www.ssig.dcc.ufmg.br/} subject to some privacy restrictions.},  contains $2,000$ images of $101$ on-track vehicles acquired at the Federal University of Minas Gerais (UFMG) campus.  Since the dataset was recorded in Brazil, the license plates have three uppercase letters, one space followed by four numbers, resulting in $14,000$ characters (alphanumeric symbols) which have been manually annotated with bounding boxes.

\begin{figure}[!t]
  \centering
  \centerline{\includegraphics[width=0.8\linewidth]{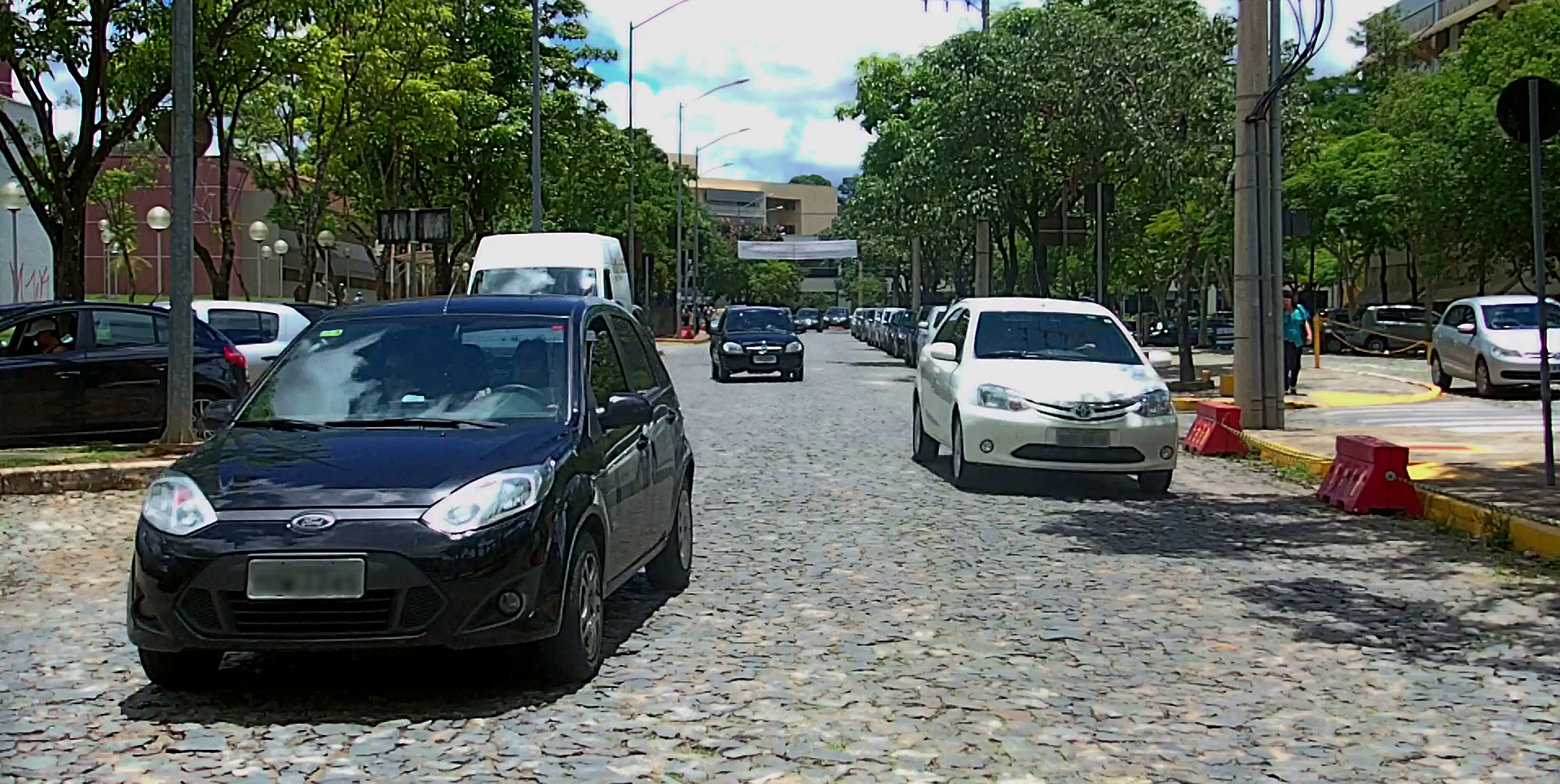}}
  \caption{Sample image of the dataset (the license plates were blurred due to privacy constraints).}
  \label{dataset/figure1} 
\end{figure}

The images of the dataset were acquired with a digital camera in Full-HD and are available in the Portable Network Graphics (PNG) format with size of $1920\times1080$ pixels. The average size of each file is $4.08$ Megabytes (a total of $8.60$ Gigabytes for the entire dataset). Figure~\ref{dataset/figure1} shows a frame sample that is present in the dataset. In addition, since there are some approaches that track the car to utilize redundant information to improve the recognition results, we decided to make a dataset with multiples frames per car (frames captured in sequence). There are, on average, $19.80$ images per vehicle (with a standard deviation of $4.14$).

A Brazilian license plate has the size of 40cm $\times$ 13cm, which means that its aspect ratio is approximately $3.08$. 
In the dataset, the license plates have sizes varying from $68\times21$~pixels to $221\times77$~pixels. On average, the license plates have the size of $120\times42$~pixels (aspect ratio of $2.86$), which is very close to actual value. In addition, each character of the Brazilian license plate has height of $6.3$~cm and the width varying according to the character. 
In the dataset, the characters in the license plate have their heights varying from $11$ to $43$~pixels, with an average of $21.19$ pixels.

\begin{figure}[!t]
  \centering
  \includegraphics[width=0.45\linewidth]{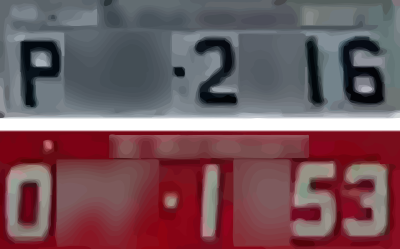}
  \caption{Example of different license plate colors in the dataset (the letters were blurred due to privacy constraints).}
  \label{dataset/figure3}
\end{figure}

\begin{figure}[!t]
  \centering
  \includegraphics[width=0.7\linewidth]{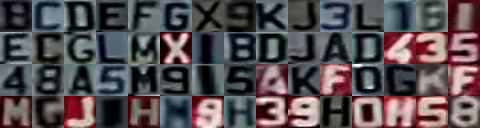}
  \caption{Example of different characters presented in our dataset.}
  \label{dataset/figure4}
\end{figure}

Our dataset is composed of images from multiple vehicle models. Among them, there are passenger vehicles ($1762$), buses and trucks ($118$), police cars ($14$) and service vehicles ($106$). This variability is important since the license plate models are not the same for all of them, as illustrated in Figure~\ref{dataset/figure3} and in Figure~\ref{dataset/figure4}. For instance, while the license plate for buses and cabs is red with white characters, it is gray with black characters for passenger vehicles. In addition, older cars might have characters of their license plates from a different text font and some license plates may be difficult to read due to the dirt that may be contained on it. Such large variance makes the proposed dataset very challenging and suitable to evaluate LPCS methods on conditions very similar to real environments.

The dataset also provide annotations regarding the position of each plate, its characters as well as the correct labeling of the plate characters, allowing a quantitative evaluation of both the plate segmentation and recognition methods. Such information is on a text file with the same name of the image.

\begin{figure*}[!t]
  \centering
  \includegraphics[width=\textwidth]{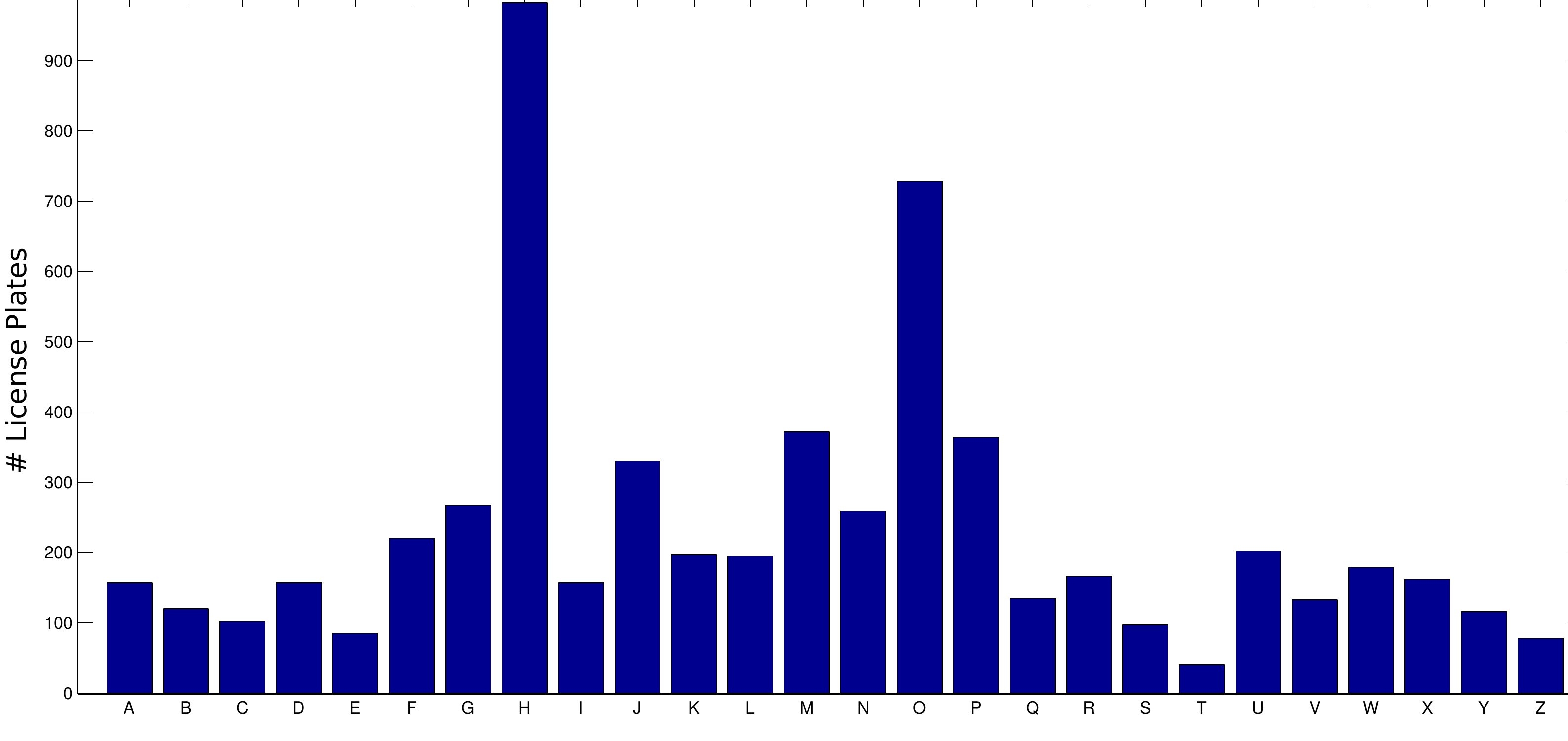}
  \caption{Frequency distribution of letters in our dataset.}
  \label{dataset/figure2}
\end{figure*}

Due to the Brazilian license plate allocation policy, the three letters of the plate are not uniformly distributed in the country. According to the State the license plate has been issued, one letter can appear much more often than others.  For instance, in Rio de Janeiro, there are more license plates with the letters K and L, in Tocantins there are more license plates with M, in Rio Grande do Sul has more with I and J, and so on. Our dataset was recorded on the State of Minas Gerais, therefore some letters are more frequent than others, as can be seen in Figure~\ref{dataset/figure2}. The letter H appears almost one thousand times, while the letters E and T occur less than one hundred times.  Although the distribution is unbalanced, we believe that it does not influence on the segmentation task because the character recognition is not being addressed at this stage of the ALPR process.

We also define a protocol to evaluate segmentation techniques. We split our dataset into three sets: training, testing and validation. Instead of using a regular division of $60\%$ of the dataset to training (model estimation), $20\%$ to validation (parameter optimization) and $20\%$ to testing (reporting final performance), we decided to provide more images for testing, resulting in the following splits: $40\%$ of the dataset to training, $20\%$ to validation and $40\%$ to testing. We keep more images on the testing set because the majority of LPCS approaches do not rely on learning techniques, i.e., do not require model estimation. This way, we are able to evaluate those methods with a large number of test images to make the reported results more statistically significant.

\begin{table*}[!t]
	\centering
    \caption{Comparison between the proposed dataset and others available in the literature, regarding different aspects. The proposed dataset is the only one to provide high resolution images with annotation of the individual characters in the license plate, essential to evaluate LPCS approaches.}
\scriptsize
    \begin{tabular}{c|cccccc}
    Dataset                                         & \multirow{2}{*}{\# Images}
                                                             & High & License Plate & Characters & Evaluation & \multirow{2}{*}{Purpose}\\ 
    ~                                               &        & Resolution & Labeled & Annotation & Protocol & \\ 
    \hline \hline
    Ferencz \& Malik~\cite{ferencz2004learning}     &  4,000 & No  & No  & No  & Yes & Car Detection               \\
    Caltech~\cite{caltech2001dataset}               &    526 & No  & No  & No  & No  & Object Recognition          \\
    VOC2006 Pascal~\cite{everingham2010pascal}      &     56 & No  & No  & No  & Yes & Object Recognition          \\
    BIT-Vehicle Dataset~\cite{dong2014vehicle}      &    900 & Yes & No  & No  & No  & Vehicle Type Classification \\
    UIUC Dataset~\cite{agarwal2004learning}         &    828 & No  & No  & No  & Yes & Vehicle Recognition         \\
    Krause Cars Dataset~\cite{krause2013collecting} & 16,185 & No  & No  & No  & Yes & Vehicle Type Classification \\
    EPFL Car Dataset~\cite{ozuysal2009pose}         &  2,000 & Yes & No  & No  & No  & Vehicle Pose Estimation     \\ \hline
    \textbf{SSIG SegPlate}                          &  2,000 & Yes & Yes & Yes & Yes & LPCS                        \\
    \end{tabular}
    \label{dataset/table1}
\end{table*}

Table~\ref{dataset/table1} provides a comparison between well-known datasets of vehicles and our proposed dataset. All of these are publicly available to the research community. The datasets have multiple purposes and do not provide neither the labels of the vehicles license plates, i.e., their  identification, e.g. ABC-1234, nor their character annotation (the bounding boxes of characters composing the license plate), essential to perform fine evaluation of LPCS methods. Some of them do not provide an evaluation protocol to their own purpose, essential to allow a fair comparison among different algorithms, and provide images in low resolution, which suggests that these images are not suitable to be employed in tasks such as LPCS. The proposed dataset overcomes the majority of these undesired characteristics found in the currently available datasets.

\subsection{Jaccard-Centroid Coefficient}
\label{subsec:measures}
Since there is no measure in the literature specifically designed to evaluate character segmentation approaches, we propose a new measure suitable to this problem, the \emph{Jaccard-Centroid} (JC) coefficient. This measure was inspired by the Jaccard coefficient, a widely employed measure to evaluate how well objects are located in images, define by
\begin{equation}
J(A, B) = \frac{A \cap B}{A \cup B},
\label{measure:J}
\end{equation}
\noindent where $A$ and $B$ are sets constituted by their bounding boxes.

There are two main motivations to create this new measure. First, the Jaccard coefficient is not very suitable to assess whether the location found by an object is well centralized according to the ground truth annotation, which is a very important feature of the segmented character for the further recognition step~\cite{menotti2014vehicle,araujo2013segmenting}. Second, to the best of our knowledge, most works in the LPCS literature do not employ a standard measure, which makes the comparison of the effectiveness of different techniques a very hard task.

\begin{figure}[!t]
  \centering
  \includegraphics[width=0.45\linewidth]{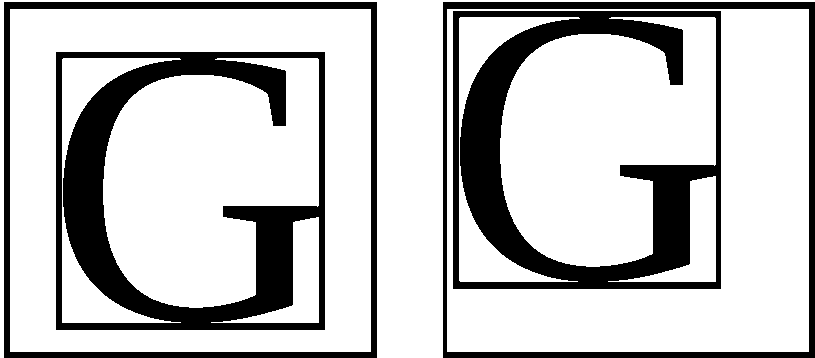}
  \caption{Illustration of two segmented bounding boxes. Both have the same Jaccard coefficient but one is not well aligned in the centroid, which might difficult the OCR step in the ALPR.}
  \label{measure:jaccardproblem}
\end{figure}

To achieve high character recognition accuracy, the segmentation task must provide characters that are easy to recognize. Menotti et al.~\cite{menotti2014vehicle} stated that a character is easily recognizable by most supervised learning techniques if the character is centralized on the bounding box. However, the Jaccard coefficient does not consider the objects alignment. For instance, Figure~\ref{measure:jaccardproblem} shows two separate bounding boxes with one smaller bounding box inside each. If we consider the inner bounding boxes as the ground truth and the outer boxes as the detection results, they have the same Jaccard coefficients. Note that we obtain the same Jaccard coefficients, even when the reverse case is considered (outer bounding boxes are the ground truth). Nonetheless, the detection on the left example is expected to be easily recognizable by an OCR since the two bounding boxes are aligned according to their center, i.e., the distance between their centroids is small. Therefore, to capture the precision of the alignment, it is necessary to combine the Jaccard coefficient and the distance between the centroids of detected and ground-truth bounding boxes, which is precisely the focus of the proposed Jaccard-Centroid coefficient.

The \emph{Jaccard-Centroid} (JC) coefficient between two bounding boxes, $JC(A,B)$, is defined as the combination of the Jaccard coefficient and the distance between the centroids of the detected and the desired objects by
\begin{equation}
JC(A, B) = \frac{J(A,B)}{\max(1,C \times \Delta c(A, B))},
\label{measure:JC}
\end{equation}
\noindent where C is a constant and $\Delta c(A, B)$ denotes the distance between the centroids of the detected and the desired objects and is defined by
\begin{equation}
 	\Delta c (A, B) = \sqrt{(A_{x}-B_{x})^2 + (A_{y}-B_{y})^2},
	\label{measure:Delta}
\end{equation}
\noindent where $(A_x,A_y)$ and $(B_x,B_y)$ represent their centroid coordinates, respectively. Note that if the centroids are perfectly aligned, the $\Delta c (A,B)$ is zero and the Jaccard-Centroid coefficient will be the same as the Jaccard coefficient.

\begin{figure}[!t]
  \centering
  \includegraphics[width=0.8\linewidth]{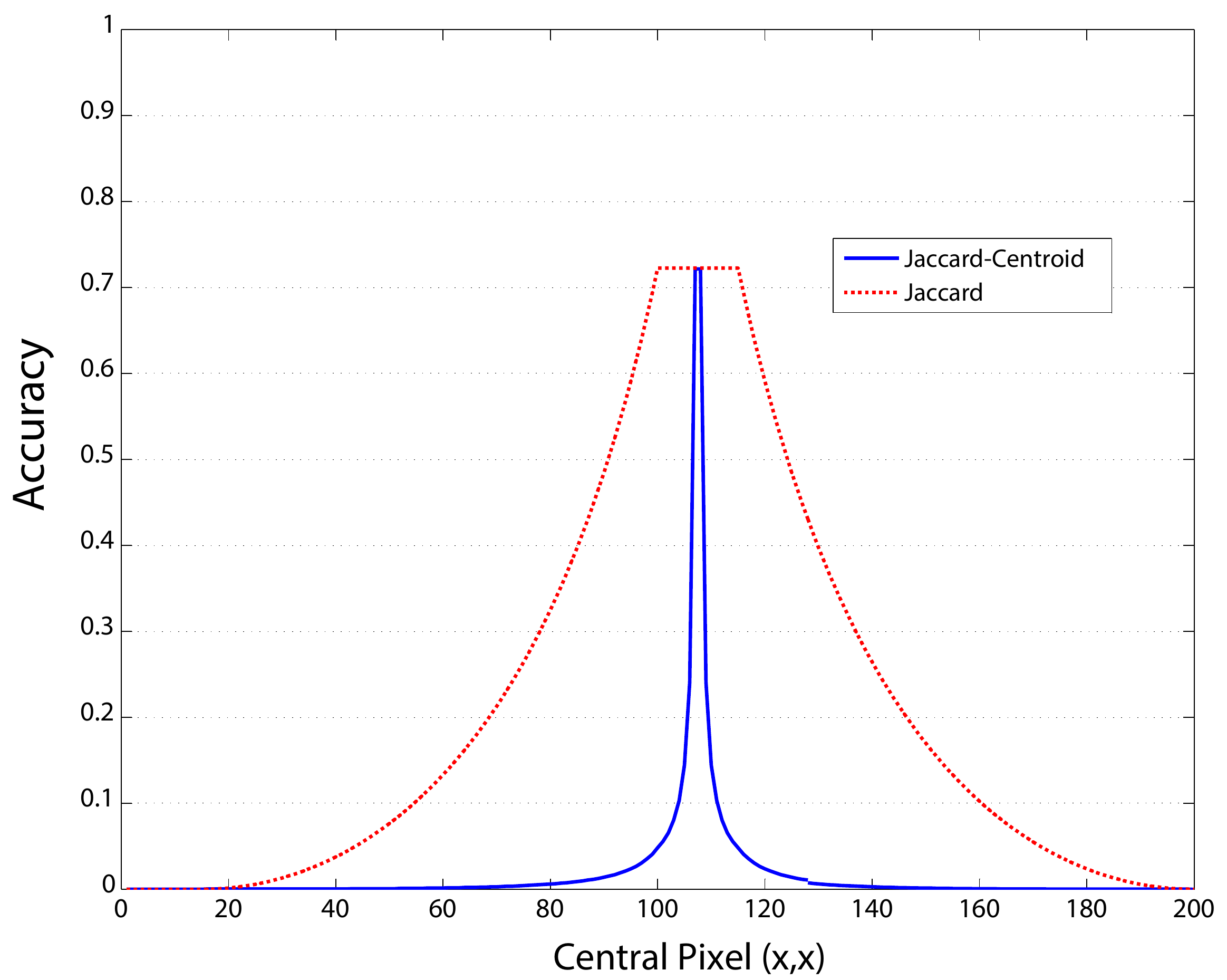}
  \caption{The graphic of the Jaccard Coefficient has a plateau when one box is completely inside the other one. However, the Jaccard-Centroid measure, with $C=2$, does not has this plateau.}
  \label{measure:jc_x_jaccard}
\end{figure}

The denominator of Equation~\ref{measure:JC} can be considered a penalty term for the Jaccard coefficient. The minimum value is 1 when the misalignment, weighted by the constant C, is less than 1. The best value for constant C was determined experimentally to maximize to recognition rate achieved by the OCR (Section~\ref{sec:OCR}).

The curves in Figure~\ref{measure:jc_x_jaccard}, obtained by sliding a window diagonally over the ground truth annotation, illustrate the difference between the Jaccard and the Jaccard-Centroid coefficients. While the curve representing the Jaccard measure has a plateau on the top (different locations lead to the same value), the Jaccard-Centroid measure presents only a peak when the centers of the bounding boxes are perfectly aligned, as desirable for performing the OCR.

\section{Experimental Results}
\label{sec:experiments}
In addition to previous mentioned baselines, in our experiments, we consider a fourth technique to segment the seven characters of the license plate using only information regarding the real shape of the Brazilian License Plate and its characters. The Brazilian license plate has seven characters and an hyphen between the third and the forth to separate letters from digits. Therefore, we consider this hyphen as one character and divide the license plate equally into eight horizontal regions. In addition, we eliminate $15\%$ of from top of the license plate and $5\%$ from the bottom to crop only the portion containing the characters.

Considering the baseline approaches described in Section~\ref{sec:related}, our proposed technique described in Section~\ref{subsec:proposed} and this fourth baseline using prior-knowledge information, we perform three main evaluations: i) individual character segmentation  (Section~\ref{sec:JCC}), ii) full license plate character segmentation to assess whether the approaches are able to segment all license plate characters(Section~\ref{sec:fullseg}); and iii)  optical character recognition on characters which were perfectly segmented and characters segmented using the baseline proposed in~\cite{nomura2009morphological} (the baseline that achieved the best accuracy in our experiments) to assess its accuracy on both scenarios (Section~\ref{sec:OCR}). We performed the experiments using the dataset described in Section~\ref{subsec:dataset}\footnote{Although we had performed experiments with a dataset containing Brazilian license plates, the proposed benchmark can be used for any type of license plates in the world with possible adaptations of the segmentation methods.}.

The segmentation and OCR approaches were implemented on the Smart Surveillance Framework~\cite{nazare2014smart} using OpenCV and C/C++ programming language. All experiments were performed on a computer with a Intel Xeon E5-2620, 32GB of RAM and a dedicated 100 GB hard drive for storage.

\subsection{Parameter Setting}

As classifier for the OCR systems, we used an One Against-All versions of the Oblique Random Forest (oRF) classifier and a SVM using a radial kernel~\cite{artur2016oblique}. As feature descriptors, we employed Histogram of Oriented Gradients (HOG)~\cite{dalal2005histograms} using 9 bins, 4 blocks and $16x16$ cell size with 50\% of stride (8 pixels) fed the classifiers.

\begin{figure}[!t]
	\centering
	\includegraphics[width=0.8\linewidth]{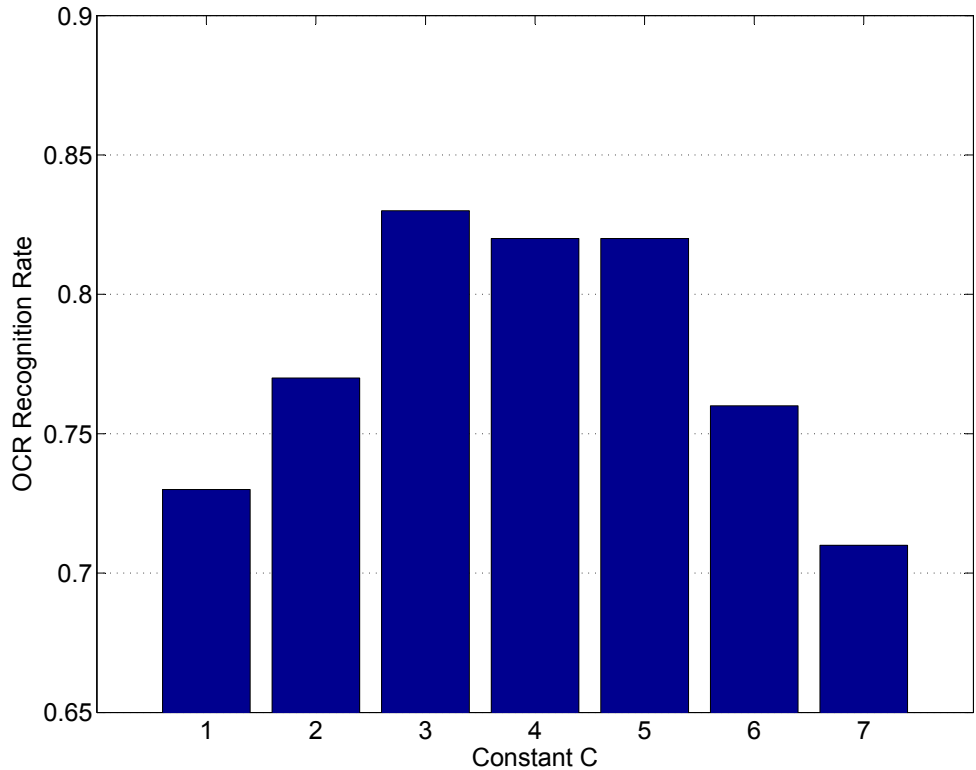}
	\caption{OCR recognition rates achieved for the first 20\%  of characters when we vary the value of the constant C from Equation~\ref{measure:JC}.}
	\label{fig:constant}
\end{figure}

To determine the best value of the constant $C$ of the Jaccard-Centroid, we executed the oRF version of the OCR on the $20\%$ best segmented characters, varying the value of $C$. 
The best achieved value was 3 as illustrated in Figure~\ref{fig:constant}. 
Therefore, all experiments reported on this paper were performed using $C=3$.

\begin{table}[!t]
    \centering
	\caption{Measure results of segmentation: average values achieved for the four baselines and our proposed approach using three measures.}
	\label{experiments:table1}
	\begin{tabular}{r|c|c|c}
		\multicolumn{1}{c|}{Approach}                                             &Jaccard&$\Delta c$&Jaccard-Centroid\\
		\hline        
		Pixel Counting with SL*L~\cite{nomura2009morphological}     &  0.561 & 2.052 & 0.316\\
        Conn. Component~\cite{shapiro2004multinational}   			&  0.452 & 1.796 & 0.235\\
		Pixel Counting with IGT~\cite{kavallieratou2006adaptive}    &  0.507 & 1.708 & 0.270\\
		Prior-Knowledge Based								    	&  0.398 & 10.820& 0.076\\
        Proposed Iterative Approach                        			&  0.601 & 1.433 & 0.419\\
	\end{tabular}
\end{table}

To the technique proposed in Shapiro \& Gluhchev~\cite{shapiro2004multinational}, we use the proportion height range of $[40\%,50\%]$ to accept a connect component as a character based on the real proportion of the height of a character regarding the height of a Brazilian license plate, which is $45\%$, as described on Section~\ref{subsec:dataset}.

\subsection{Individual Character Segmentation Evaluation}
\label{sec:JCC}

Table~\ref{experiments:table1} shows the average values achieved by the four segmentation methods on the proposed dataset.
On one hand, the segmentation by the Prior Knowledge-Based approach (expected due to its simplicity) presents a higher average degree of misalignment, represented by the $\Delta c$. 
As a consequence, this segmentation approach is penalized by the proposed Jaccard-Centroid measure (its value is $0.322$ lower than the value computed using the Jaccard coefficient). 
Therefore, the accuracy of the OCR using the characters segmented by the Prior Knowledge-Based is expected to be reduced due to this misalignment. On the other hand, the connected component labeling and both pixel counting approaches achieved smaller $\Delta c$ value, causing minor penalization to the Jaccard-Centroid coefficient. The SL*L using Pixel Counting was capable to achieve an average scores near $0.56$ by Jacard measure and near $0.30$ by Jaccard-Centroid coefficient, which is the best result of the four proposed baseline.

Our proposed approach was the best evaluated one. It achieved a higher value in Jaccard and it is not much penalized by the $\Delta$ value, which corresponds to low misalignment error. These results supports the hypothesis that our method, despite being straightforward, is the best approach to perform LPCS efficiently.

\begin{figure}[!t]
	\centering
	\includegraphics[width=0.8\linewidth]{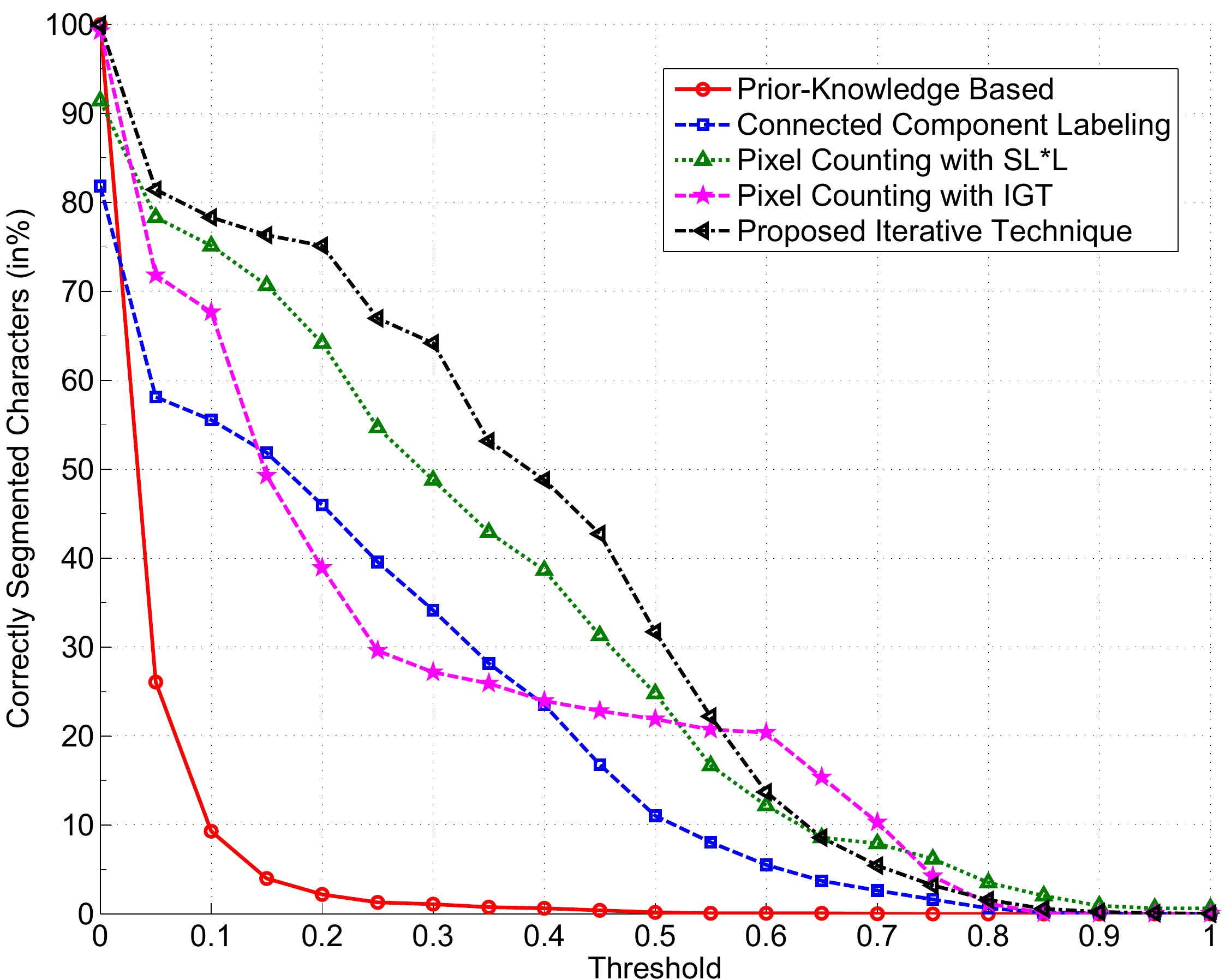}
	\caption{Percentage of individual characters correctly segmented as a function of the Jaccard-Centroid coefficient.}
	\label{experiments/figure2}
\end{figure}

We also analyzed the number of characters that were satisfactory segmented as a function of the Jaccard-Centroid coefficient. Figure~\ref{experiments/figure2} shows curves of the effectiveness (the correctly segmented characters) of our proposed method and each evaluated baseline approach as a function of the threshold on the Jaccard-Centroid measure. That is, for a given threshold value (ranging from 0.05 to 1), we compute the percentage of characters that have obtained a Jaccard-Centroid measure equal or higher than the threshold.

For an analysis, we consider $0.4$ as a Jaccard-Centroid threshold to obtain a satisfactory segmentation, otherwise, the character might not be well-centered and the OCR will not work properly. According to the results, neither of the baselines approaches is accurate enough to be employed in a reliable ALPR system. The approaches using SL*L and IGT followed by Pixel Counting were capable of segmenting satisfactorily around $25\%$ and $40\%$ of the characters, while the technique using connected component labeling and the approach using Prior-Knowledge were able to segment only $25\%$ and $3\%$, respectively. Our approach was capable to segment around $50\%$ of all license plate characters, achieving the best results among the evaluated methods.

\subsection{Full License Plate Segmentation Evaluation}
\label{sec:fullseg}

In this section, we evaluate the segmentation of the entire license plate to analyze its relation with the Jaccard-Centroid coefficient. We used the values of our measure applied to the seven characters of one license plate to determine whether the license plate segmentation would be plausible for recognition or not. This is an important evaluation since all characters must be found in the plate and each of them must be well located/segmented so that the plate can be properly recognized by OCR techniques.

\begin{figure}[!tb]
	\centering
	\includegraphics[width=0.8\linewidth]{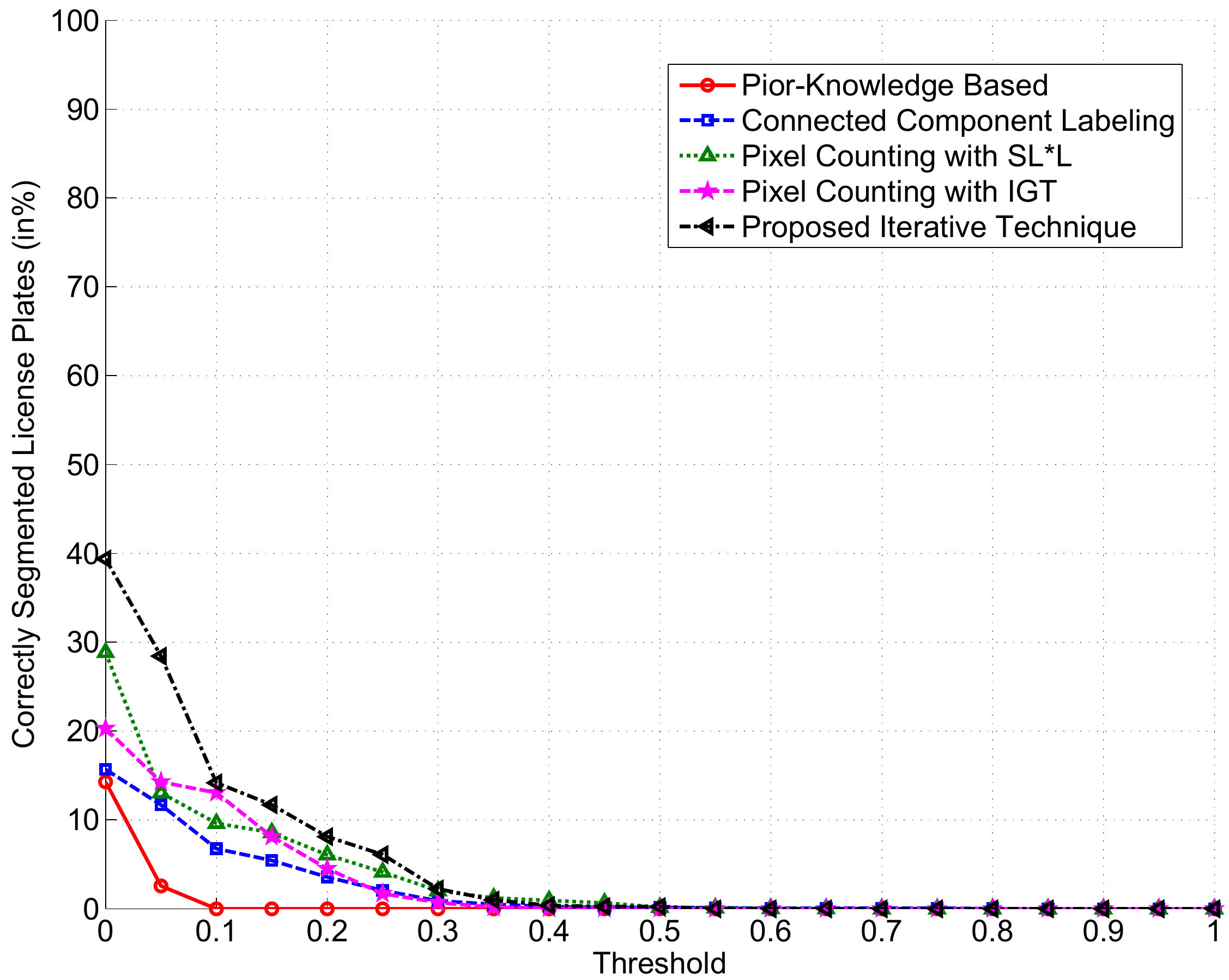}
	\caption{Percentage of correctly segmented license plates (all seven characters were segmented in the plate) as a function of the Jaccard-Centroid coefficient. }
	\label{experiments/figure1}
\end{figure}

To perform the evaluation, we analyze the Jaccard-Centroid coefficient values by varying a threshold. That is, if the character of the license plate with the lowest Jaccard-Centroid coefficient is higher or equal than the current threshold, the license plate is considered as correctly segmented. Whenever an approach finds out a number of characters different of the one expected, we consider that the JC of the license plate is 0, that is, the license plate is not correctly segmented at all. Figure~\ref{experiments/figure1} shows how the correctly segmented license plates percentage varies as a function of the Jaccard-centroid coefficient for each approach. According to the results, none of the approaches presents high accuracy for higher threshold values.

Considering the same threshold used in the previous section ($0.4$), the prior knowledge based approach is not able to segment any license plate, confirming that this method is too simple to perform LPCS. Besides, the connected component labeling approach and IGT with pixel counting approach achieved segmentation rates close to $2\%$ and $3\%$, respectively on the mentioned threshold which also entails a very low performance. The approach using SL*L and pixel counting achieved $6\%$ segmentation rate, which would not be considered satisfactory for any real application. Our approach was capable to achieve $8\%$ of segmentation rate which, despite being the best result of all evaluated techniques, it is also not satisfactory enough to be employed in real scenarios. Such results reinforce the fact that our dataset is challenging and suitable to evaluate the robustness of LPCS techniques.

According to the results showed in Figures~\ref{experiments/figure2} and~\ref{experiments/figure1}, it is possible to see that, even though the approach is capable of segmenting $50\%$ of the characters, it is not able to segment all characters in more than $8\%$ of license plates. This fact shows that in almost all license plates, there is at least one character that is not well segmented by any approach, which is critical for the license plate recognition once the OCR requires an acceptable segmentation for correctly recognize all characters -- if a single character was not well-segmented, the identification of the license plate is compromised.

\begin{figure}[!t]
	\centering
	\includegraphics[width=\linewidth]{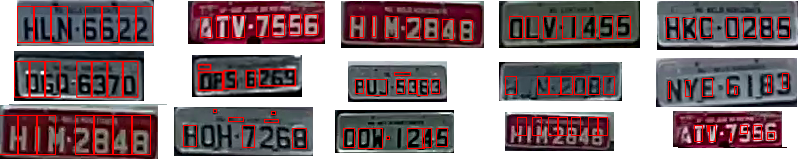}
	\caption{Examples of the segmentations by each baseline. First column shows license plates segmented by the Prior-Knowledge approach. Second column shows license plates segmented by CCL approach. Third column by SLL approach. Fourth column by IGT approach. Fifth column by the proposed iterative approach.}
	\label{experiments/figure4}
\end{figure}

Figure~\ref{experiments/figure4} shows examples of license plates segmented by each one of the approaches evaluated in this benchmark. Note that the first license plate of each column was a considered good segmentation for the method. The other two license plates of each column present explicit segmentation problems.

\subsection{Optical Character Recognition Evaluation}
\label{sec:OCR}

An accurate segmentation is crucial to an ALPR system once a poor segmentation can lead to a low accuracy by the OCR method. To support that, we performed experiments to evaluate the accuracy of the OCR when applied to license plate characters segmented with and without a precise segmentation.

\begin{table}[!t]
	\centering
	\caption{Recognition rates of the OCR using both segmentation approaches (manual and automatic) and two classifiers (Radial SVM and oRF).}
	\label{tab:perfectSeg}
	\begin{tabular}{r|cc|cc}
		  ~				    &\multicolumn{2}{c|}{Manual}&\multicolumn{2}{c}{Segmentation by}\\
		Approach		    &\multicolumn{2}{c|}{Segmentation}&\multicolumn{2}{c}{Pixel Counting}\\
		\hline
			~			   & Letters & Numbers & Letters & Numbers\\
		Radial SVM		   &  0.919 & 0.962 & 0.275 & 0.552\\
		oRF				   &  0.947 & 0.969 & 0.586 & 0.782\\
	\end{tabular}
\end{table}

Table~\ref{tab:perfectSeg} demonstrates the accuracy of the two mentioned learning based OCR systems obtained when applied to: (i) manually segmented characters and (ii) automatically segmented characters by~\cite{nomura2009morphological} (Pixel Counting approach). According to the results, there is a large influence of the character segmentation on the final results of the OCR. For instance, the OCR recognition rate can decrease on 0.644 and 0.410 points in the worst-case, for letters and numbers respectively, justifying therefore the need to have a precise segmentation system in the ALPR pipeline. It is worth to point out that spatial invariant is closely tied to the OCR implementation. Therefore, some OCR techniques can handle few misplacement better than the ones used in this work. Nonetheless, misplacement problem does not happen only when the character is uncentralized, it can also occurs when the character boundaries are outside the bounding box. Hence, spatial invariant approaches such convolutional neural networks are also expect to have their performance diminished when with poor segmented characters.

\begin{figure}[!t]
	\centering
	\includegraphics[width=0.8\linewidth]{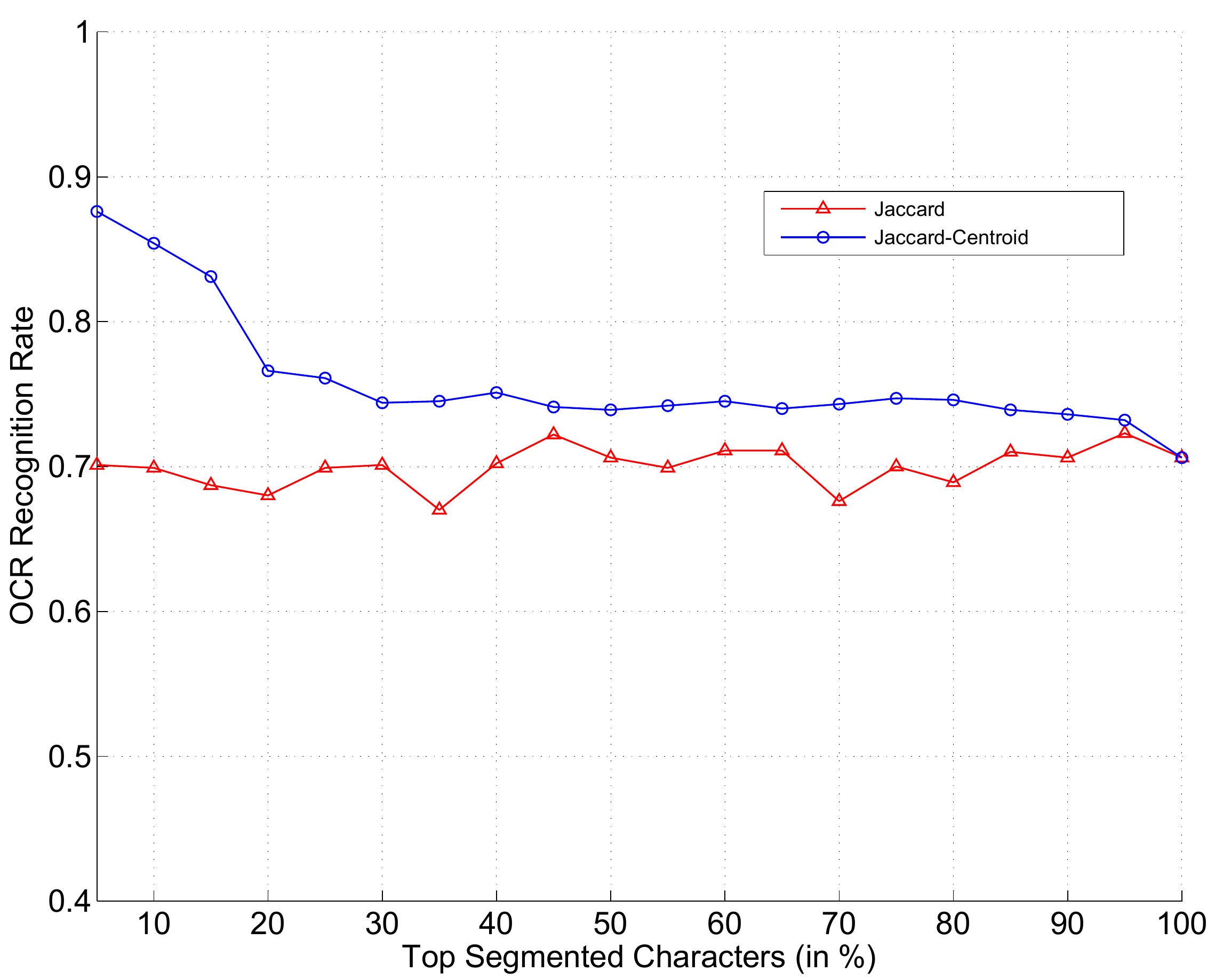}
	\caption{Recognition rate of OCR as a function of a percentage of the top segmented characters considering Jaccard and Jaccard-Centroid coefficients.}
	\label{experiments/figure3}
\end{figure}

Finally, given a rank of the characters that are best segmented according to Jaccard and Jaccard-Centroid coefficients, Figure~\ref{experiments/figure3} shows the recognition rates of an OCR system when applied to a percentage of the top segmented characters of these ranks. The x-axis represents the proportion of the top characters that were evaluated and the y-axis represents the OCR recognition rate. According to the results, using $5\%$ best segmented characters, the Jaccard-Centroid achieves an OCR recognition rate is $10\%$ higher than the one of  Jaccard coefficient. This demonstrates that the proposed Jaccard-Centroid coefficient can assign high values to characters that are easier to be recognized by an OCR, differently from the original Jaccard measure. 

\section{Conclusions}
\label{sec:conclusion}
This paper introduced a new benchmark to the license plate character segmentation (LPCS) problem.  This benchmark includes a new dataset with $2,000$ images of $101$ different on-road vehicles, spanning a total of $14,000$ alphanumerical symbols (letters and numbers), and a new measure to evaluate the effectiveness of character segmentation approaches called Jaccard-Centroid.  As a consequence of the use of our measure, the benchmark also includes curves built by varying an threshold on Jaccard-Centroid coefficient and analyzing the full correctly segmented license plates with all characters over that threshold.

We also evaluated our proposed character segmentation technique and four LPCS approaches as baselines and computed their score on the new dataset. The best result was achieved by our proposed iterative approach. The results demonstrated that the new dataset is very challenging since none of the implemented approaches achieved average values above $0.32$ (in a range between $0$ and $1$) according to the new measure. Furthermore, if we consider $0.4$ as a satisfactory Jaccard-Centroid threshold to determine whether the characters in the plate were correctly segmented (from our experience with OCR, near or perfect recognition accuracy can be achieved only when Jaccard-Centroid is equal or greater than $0.4$), none of the approaches was capable of segmenting all characters in more than $10\%$ of the license plates.

As future works, we intend to collect more images to create an extension of the dataset SSIG-SegPlate1 with more than $10,000$ images of on-road vehicles and at least $1,000$ samples of each character to perform an extensive analysis of OCR techniques in the ALPR context. We also intend to incorporate license plate images that utilizes other alphabet characters than the one used in Brazil.


\balance
\bibliography{paper}

\begin{thebibliography}{10}

\bibitem{du2013automatic}
S.~Du, M.~Ibrahim, M.~Shehata, and W.~Badawy, ``Automatic license plate
  recognition ({ALPR}): A state-of-the-art review,'' {\em Circuits and Systems
  for Video Technology, IEEE Trans. on} {\bf 23}(2), 311--325  (2013).

\bibitem{prates2014brazilian}
R.~F. Prates, G.~C{\'a}mara-Ch{\'a}vez, W.~R. Schwartz, and D.~Menotti,
  ``Brazilian {L}icense {P}late detection using histogram of oriented gradients
  and sliding windows,'' {\em International Journal of Computer Science \&
  Information Technology (IJCSIT)} {\bf 5}(6), 39--52  (2013).

\bibitem{mendes2011}
P.~R. Mendes, J.~M.~R. Neves, A.~I. Tavares, and D.~Menotti, ``Towards an
  automatic vehicle access control system: License plate location,'' in {\em
  Systems, Man, and Cybernetics (SMC), 2011 IEEE International Conf. on},
  2916--2921  (2011).

\bibitem{sarfraz2003saudi}
M.~Sarfraz, M.~J. Ahmed, and S.~A. Ghazi, ``Saudi arabian license plate
  recognition system,'' in {\em Geometric Modeling and Graphics, Int. Conf.
  on},  36--41  (2003).

\bibitem{menotti2014vehicle}
D.~Menotti, G.~Chiachia, A.~X. Falcao, and V.~J. {Oliveira Neto}, ``Vehicle
  license plate recognition with random convolutional networks,'' in {\em
  Graphics, Patterns and Images, SIBGRAPI Conf.},  298--303  (2014).

\bibitem{araujo2013segmenting}
L.~Ara{\'u}jo, S.~Pio, and D.~Menotti, ``Segmenting and recognizing license
  plate characters,'' in {\em Workshop of Undegraduate Works - Graphics,
  Patterns and Images, SIBGRAPI Conf. on},  251--270  (2013).

\bibitem{nomura2009morphological}
S.~Nomura, K.~Yamanaka, T.~Shiose, H.~Kawakami, and O.~Katai, ``Morphological
  preprocessing method to thresholding degraded word images,'' {\em Pattern
  Recognition Letters} {\bf 30}(8), 729--744  (2009).

\bibitem{shapiro2004multinational}
V.~Shapiro and G.~Gluhchev, ``Multinational license plate recognition system:
  Segmentation and classification,'' in {\em Pattern Recognition, (ICPR)
  International Conf. on},   {\bf 4}, 352--355, IEEE  (2004).

\bibitem{kavallieratou2006adaptive}
E.~Kavallieratou and S.~Stathis, ``Adaptive binarization of historical document
  images,'' in {\em 18th International Conference on Pattern Recognition
  (ICPR'06)},   {\bf 3}, 742--745, IEEE  (2006).

\bibitem{anagnostopoulos2008license}
C.-N. Anagnostopoulos, I.~E. Anagnostopoulos, I.~D. Psoroulas, V.~Loumos, and
  E.~Kayafas, ``License plate recognition from still images and video
  sequences: A survey,'' {\em Intelligent Transportation Systems, IEEE Trans.
  on} {\bf 9}(3), 377--391  (2008).

\bibitem{tan2012new}
J.~Tan, J.-H. Lai, C.-D. Wang, W.-X. Wang, and X.-X. Zuo, ``A new handwritten
  character segmentation method based on nonlinear clustering,'' {\em
  Neurocomputing} {\bf 89}, 213--219  (2012).

\bibitem{ciresan2011convolutional}
D.~C. Ciresan, U.~Meier, L.~M. Gambardella, and J.~Schmidhuber, ``Convolutional
  neural network committees for handwritten character classification,'' in {\em
  Document Analysis and Recognition (ICDAR), 2011 International Conf. on},
  1135--1139, IEEE  (2011).

\bibitem{roy2012multi}
P.~P. Roy, U.~Pal, J.~Llad{\'o}s, and M.~Delalandre, ``Multi-oriented touching
  text character segmentation in graphical documents using dynamic
  programming,'' {\em Pattern Recognition} {\bf 45}(5), 1972--1983  (2012).

\bibitem{neumann2012real}
L.~Neumann and J.~Matas, ``Real-time scene text localization and recognition,''
  in {\em Computer Vision and Pattern Recognition (CVPR), 2012 IEEE Conf. on},
  3538--3545, IEEE  (2012).

\bibitem{wang2013novel}
R.~Wang, G.~Wang, J.~Liu, and J.~Tian, ``A novel approach for segmentation of
  touching characters on the license plate,'' in {\em 2012 International Conf.
  on Graphic and Image Processing},  876847--876847, International Society for
  Optics and Photonics  (2013).

\bibitem{kahraman2003license}
F.~Kahraman, B.~Kurt, and M.~G{\"o}kmen, ``License plate character segmentation
  based on the gabor transform and vector quantization,'' in {\em Computer and
  Information Sciences - ISCIS 2003: 18th International Symposium},
  A.~Yaz{\i}c{\i} and C.~{\c{S}}ener, Eds., 381--388, Springer Berlin
  Heidelberg  (2003).

\bibitem{xia2011study}
H.~Xia and D.~Liao, ``The study of license plate character segmentation
  algorithm based on vertical projection,'' in {\em Consumer Electronics,
  Communications and Networks (CECNet), 2011 International Conf. on},
  4583--4586, IEEE  (2011).

\bibitem{jagannathan2013license}
J.~Jagannathan, A.~Sherajdheen, R.~Deepak, and N.~Krishnan, ``License plate
  character segmentation using horizontal and vertical projection with dynamic
  thresholding,'' in {\em Emerging Trends in Computing, Communication and
  Nanotechnology (ICE-CCN), 2013 International Conf. on},  700--705, IEEE
  (2013).

\bibitem{soumya2014license}
K.~R. Soumya, A.~Babu, and L.~Therattil, ``License plate detection and
  character recognition using contour analysis,'' {\em Int. Journal of Advanced
  Trends in Computer Science and Engineering} {\bf 3}(1), 15--18  (2014).

\bibitem{xing2012new}
F.~Xing-lin and F.~Yun-lou, ``A new license plate character segmentation
  algorithm based on priori knowledge constraints,'' {\em Journal of Chongqing
  Technology and Business University (Natural Science Edition)} {\bf 8}, 011
  (2012).

\bibitem{chuang2014vehicle}
C.-H. Chuang, L.-W. Tsai, M.-S. Deng, J.-W. Hsieh, and K.-C. Fan, ``Vehicle
  licence plate recognition using super-resolution technique,'' in {\em
  Advanced Video and Signal Based Surveillance (AVSS), IEEE Int. Conf. on},
  411--416  (2014).

\bibitem{fan2012license}
Z.~Fan, Y.~Zhao, A.~M. Burry, and V.~Kozitsky, ``License plate character
  segmentation using likelihood maximization,''  (2012).
\newblock {US} Patent App. 13/464,357, Google Patents.

\bibitem{franc2005license}
V.~Franc and V.~Hlav{\'a}{\v{c}}, ``License plate character segmentation using
  hidden markov chains,'' in {\em Pattern Recognition: 27th DAGM Symposium},
  W.~G. Kropatsch, R.~Sablatnig, and A.~Hanbury, Eds., 385--392, Springer
  Berlin Heidelberg  (2005).

\bibitem{nagare2011license}
A.~P. Nagare, ``License plate character recognition system using neural
  network,'' {\em International Journal of Computer Applications} {\bf 25}(10),
  36--39  (2011).

\bibitem{guo2008license}
J.-M. Guo and Y.-F. Liu, ``License plate localization and character
  segmentation with feedback self-learning and hybrid binarization
  techniques,'' {\em Vehicular Technology, IEEE Trans. on} {\bf 57}(3),
  1417--1424  (2008).

\bibitem{antonacopoulos2009realistic}
A.~Antonacopoulos, D.~Bridson, C.~Papadopoulos, and S.~Pletschacher, ``A
  realistic dataset for performance evaluation of document layout analysis,''
  in {\em Document Analysis and Recognition, 2009. ICDAR'09. 10th International
  Conf. on},  296--300, IEEE  (2009).

\bibitem{guyon1994unipen}
I.~Guyon, L.~Schomaker, R.~Plamondon, M.~Liberman, and S.~Janet, ``Unipen
  project of on-line data exchange and recognizer benchmarks,'' in {\em Pattern
  Recognition, Computer Vision \& Image Processing., IAPR International. Conf.
  on},   {\bf 2}, 29--33  (1994).

\bibitem{yao2012detecting}
C.~Yao, X.~Bai, W.~Liu, Y.~Ma, and Z.~Tu, ``Detecting texts of arbitrary
  orientations in natural images,'' in {\em Computer Vision and Pattern
  Recognition (CVPR), 2012 IEEE Conf. on},  1083--1090, IEEE  (2012).

\bibitem{agarwal2002learning}
S.~Agarwal and D.~Roth, ``Learning a sparse representation for object
  detection,'' in {\em 7th European Conference on Computer Vision-Part IV},
  {\em ECCV '02}, 113--130, Springer-Verlag  (2002).

\bibitem{agarwal2004learning}
S.~Agarwal, A.~Awan, and D.~Roth, ``Learning to detect objects in images via a
  sparse, part-based representation,'' {\em Pattern Analysis and Machine
  Intelligence, IEEE Trans. on} {\bf 26}(11), 1475--1490  (2004).

\bibitem{caltech2001dataset}
``Caltech object category datasets..''
  \url{http://www.vision.caltech.edu/archive.html}  (2001).
\newblock {A}ccessed on: 2015-03-25.

\bibitem{krause2013collecting}
J.~Krause, J.~Deng, M.~Stark, and F.-F. Li, ``Collecting a large-scale dataset
  of fine-grained cars.''
  https://www.d2.mpi-inf.mpg.de/sites/default/files/fgvc13.pdf  (2013).

\bibitem{dong2014vehicle}
Z.~Dong, M.~Pei, Y.~He, T.~Liu, Y.~Dong, and Y.~Jia, ``Vehicle type
  classification using unsupervised convolutional neural network,'' in {\em
  Pattern Recognition (ICPR), 2014 22nd International Conference on},
  172--177, IEEE  (2014).

\bibitem{ferencz2004learning}
A.~D. Ferencz, E.~G. Learned-Miller, and J.~Malik, ``Learning hyper-features
  for visual identification,'' in {\em Advances in Neural Information
  Processing Systems},  425--432  (2004).

\bibitem{ozuysal2009pose}
M.~Ozuysal, V.~Lepetit, and P.~Fua, ``Pose estimation for category specific
  multiview object localization,'' in {\em Computer Vision and Pattern
  Recognition, 2009. CVPR 2009. IEEE Conference on},  778--785, IEEE  (2009).

\bibitem{otsu1975threshold}
N.~Otsu, ``A threshold selection method from gray-level histograms,'' {\em IEEE
  Trans. on Systems, Man, and Cybernetics} {\bf 9}(9), 62--66  (1979).

\bibitem{serra1986introduction}
J.~Serra, ``Introduction to mathematical morphology,'' {\em Computer vision,
  graphics, and image processing} {\bf 35}(3), 283--305  (1986).

\bibitem{kavallieratou2005binarization}
E.~Kavallieratou, ``A binarization algorithm specialized on document images and
  photos.,'' in {\em ICDAR},   {\bf 5}, 463--467  (2005).

\bibitem{matas2005unconstrained}
J.~Matas and K.~Zimmermann, ``Unconstrained licence plate and text localization
  and recognition,'' in {\em Intelligent Transportation Systems, 2005.
  Proceedings. 2005 IEEE},  225--230, IEEE  (2005).

\bibitem{gonzalez2009digital}
R.~C. Gonzalez, {\em Digital image processing}, Pearson Education India
  (2009).

\bibitem{everingham2010pascal}
M.~Everingham, L.~Van~Gool, C.~K. Williams, J.~Winn, and A.~Zisserman, ``The
  pascal visual object classes (voc) challenge,'' {\em International journal of
  computer vision} {\bf 88}(2), 303--338  (2010).

\bibitem{nazare2014smart}
A.~C. {Nazare Jr.}, C.~E. dos Santos, R.~Ferreira, and W.~R. Schwartz, ``{Smart
  Surveillance Framework: A Versatile Tool for Video Analysis},'' in {\em IEEE
  Winter Conf. on Applications of Computer Vision (WACV)},  753--760  (2014).

\bibitem{artur2016oblique}
A.~Jordao and W.~R. Schwartz, ``Oblique random forest based on partial least
  squares applied to pedestrian detection,'' in {\em IEEE International
  Conference on Image Processing - ICIP},   ((accepted)2016).

\bibitem{dalal2005histograms}
N.~Dalal and B.~Triggs, ``Histograms of oriented gradients for human
  detection,'' in {\em Computer Vision and Pattern Recognition, IEEE Conf. on},
    {\bf 1}, 886--893  (2005).

\end{thebibliography}
\bibliographystyle{spiejour}   


\vspace{2ex}\noindent\textbf{Gabriel Resende Gonçalves} received a bachelor degree in Computer Science from Universidade Federal de Ouro Preto, Brazil and a master degree in Computer Science from Universidade Federal de Minas Gerais, Brazil. Currently, he is working as a junior researcher at Foundation from Reseach Development. Has experience in Pattern Recognition, Data Mining and Information Retrieval.

\vspace{2ex}\noindent\textbf{Sirlene Pio Gomes da Silva} received a Bachelor degree and a master degree in Computer Science both from Universidade Federal de Ouro Preto, Brazil. Has experience in Deep Learning applied to Pattern Recognition, Computer Vision and Image Processing.

\vspace{2ex}\noindent\textbf{David Menotti} is an Associate Professor with the Department of Informatics from Universidade Federal do Paraná, Brazil. He received the master’s degrees in computer engineering and applied informatics from the Pontifícia Universidade Católica do Paraná, Brazil, in 2001 and 2003, respectively, and the Ph.D. degree in computer science from the Universidade Federal de Minas Gerais Brazil in 2008.  His research interests include machine learning, image processing, pattern recognition, computer vision, and information retrieval.

\vspace{2ex}\noindent\textbf{William Robson Schwartz} is an Associate Professor in the Department of Computer Science at the Universidade Federal de Minas Gerais, Brazil. Received PhD degree from University of Maryland, College Park, USA. His research interests include Computer Vision, Smart Surveillance, Forensics, and Biometrics, in which he authored more than 100 scientific papers and coordinates projects sponsored by several Brazilian Funding Agencies. He is also the head of the Smart Surveillance Interest Group (SSIG).

\listoffigures
\listoftables

\end{spacing}
\end{document}